# Through the Twitter Glass: Detecting Questions in Micro-Text

**Kyle Dent** and **Sharoda Paul**
kdent@parc.com and spaul@parc.com
Palo Alto Research Center
3333 Coyote Hill Road
Palo Alto, CA 94304

## Abstract

In a separate study, we were interested in understanding people's Q&A habits on Twitter. Finding questions within Twitter turned out to be a difficult challenge, so we considered applying some traditional NLP approaches to the problem. On the one hand, Twitter is full of idiosyncrasies, which make processing it difficult. On the other it is very restricted in length and tends to employ simple syntactic constructions, which could help the performance of NLP processing. In order to find out the viability of NLP and Twitter, we built a pipeline of tools to work specifically with Twitter input for the task of finding questions in tweets. This work is still preliminary, but in this paper we discuss the techniques we used and the lessons we learned.

## Introduction

If you try to sip from the fire hose of Twitter, you might feel like Alice at the tea party in 'Through the Looking-Glass,' a swirling confusion of short personal remarks and people asking impossible riddles. You get the sense that extracting meaningful information from the stream of seeming nonsense verse presents a difficult task, and even if it were possible, the chaff of Twitter feels like empty casing. However, there are ripe seeds and bits of useful information. Apart from being used by people to share daily chatter (Java et al. 2007), Twitter has been used for communication during mass emergencies (Corvey et al. 2010), for informal communication in organizations (Zhao and Rosson 2009), and even as a platform for online activism (Gaffney 2010). The vast amount of Twitter data being generated every day provides exciting opportunities for analysis.

Indeed, we have been interested in understanding the extent to which people use Twitter in seeking information and recommendations. We launched a study whose goal was to explore how Twitter is used by people for social question-answering (Q&A) (S.A. Paul and Chi 2011a). Recent studies have found that people often turn to their online social networks to fulfill their information needs (Morris, Teevan, and Panovich 2010). People turn to their friends on social networks to ask subjective questions which are hard to answer using search engines. People trust their friends to provide tailored, contextual responses; hence, social network-based Q&A is becoming popular (S.A. Paul and Chi 2011b). We wanted to study the types and topics of questions being asked on Twitter, the responses received to those questions, and the effect of the underlying social network on this Q&A behavior. The first step in our study was to identify question tweets. Detecting questions in online content is challenging and remains an active area of research (Wang and Chua 2010). Detecting questions on Twitter is especially difficult due to the additional challenges posed by the short, informal nature of tweets. We were, therefore, interested in exploring NLP techniques that might help us identify question tweets. We might have tried some simpler approaches such as a binary classifier for finding questions, but part of the exercise was to get a sense of the difficulty of understanding the language of Twitter, It would also be interesting to determine a baseline using machine learning classifiers for comparison. This is something we hope to do in the future.

## NLP and Micro-text

Micro-text has several characteristics that distinguish it from traditional documents (Rosa and Ellen 2009). First, micro-text is short, often consisting of a single sentence or even a single word. Second, the grammar used is informal and unstructured, and there are often abbreviations and errors. (Rosa and Ellen 2009). Twitter is a popular micro-blogging service where users post updates, or tweets, that are up to 140 characters long. The unique characteristics of tweets make it challenging to apply standard NLP tools to analyze this data. For instance, Named Entity Recognizers have been shown to perform extremely poorly on tweets (Kaufmann 2010).

NLP in general has not been wholly successful in getting computers to the point of understanding all of human language, or even a single human language. While Twitter seems to be full of inconsistent and highly creative use of language is it possible that the enforced shortness of input strings in Twitter can make language processing easier? Are there qualities of the language used in Twitter that make the problem less complicated? Based on the assumption that this is true, we set out to create a traditional NLP parser for tweets to tackle the question of questions in Twitter.

Believing that a valid syntactic question might be a strong indicator of an information-seeking tweet, we developed a linguistic parser to search through tweets to decide if that is the case. In the end we determined that by itself well-formedness is not a strong enough indicator to differentiate between actual questions and tweets that are formed as questions but are asked either rhetorically or to convey some other sense. Consider the following tweets.

(1) In UK when you need to see a specialist do you need special forms or permission?

(2) What the chances of both of my phones going off at the same time?? Smh in a minor meeting... Ugh!

(3) How has everyone's day gone so far? Today is going too fast for me!

Example 1 is undoubtedly looking for an answer to a specific question, while the second is meant to express surprise and would not expect an answer. The third follows a common pattern where the question is presented as a poll of followers, but most likely is only meant to introduce a topic that the tweeter wants to respond to herself.

Tweets that are not presented as questions at all, might also be looking for information, and in our case, we still want to know about them. In the end, simply understanding the syntax is not interesting enough. Invariably, there are more profound issues of concern, and syntax by itself does not get you far enough to understand them.

## NLP Challenges

Issues in implementing NLP approaches to understanding text in practical systems are well understood. We discuss a few of them here and suggest that in Twitter text, they are less of a problem.

Natural human languages are mostly context-free equivalent. Parsing with a context-free grammar is for the most part a cubic operation, and in the face of input such as Example 4 most parsers will fail.

(4) The trunks being now ready, he, after kissing his mother and sisters, and once more pressing to his bosom his adored Gretchen, who, dressed in simple white muslin, with a single tuberose in the ample folds of her rich brown hair, had tottered feebly down the stairs, still pale from the terror and excitement of the past evening, but longing to lay her poor aching head yet once again upon the breast of him whom she loved more dearly than life itself, departed.[1]

At best a parser takes several minutes or more to finish although more likely it will exhaust available memory without ever finishing. Given the 140-character limit of Twitter, parsers never have to cope with input this long or with difficult embedding or long-distance dependencies.

There are other well-established difficulties in understanding language, for example attachment ambiguity. With a sentence like:

(5) The butler hit the intruder with a hammer.

---

[1] Thanks to Mark Twain and his essay "The Awful German Language." Although he contrived this text (to make a similar point in a different context), it is not difficult to find equally dense examples capable of choking many parsers.

Most speakers will understand that the butler used a hammer to hit the intruder; however, both of the following parse trees are possible.

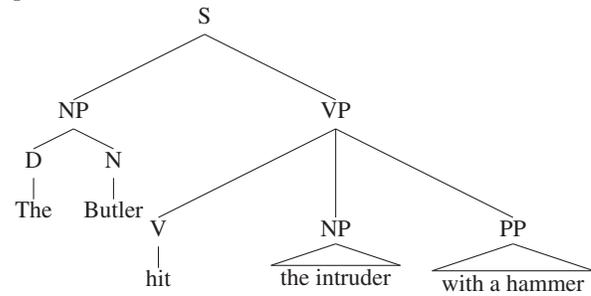

The PP can attach to the VP as the instrument of the butler's assault.

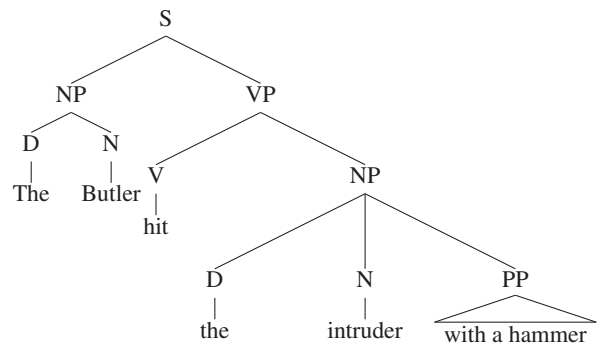

Or to the intruder describing him.

Complicated coordination is another common problem.

(6) The patient was asked to bring her pathology slides for review here at the clinic and also an updated report.

In this example, *and* conjoins *her pathology slides* and *an updated report* although it would be difficult for a parser to determine that *the clinic* is not the first conjunct.

Notice that in the last two examples, the sentences are within the 140-character limit. The short length of the text does not provide any advantage. However, in analyzing Twitter data, we have noticed that sentences tend to follow simpler constructions and that many of the ambiguity problems do not exist or exist to a much smaller extent. We believe that utterances in Twitter follow their own rules of a grammar, which is certainly parsable by humans and quite possibly by machines as well. At the very least, the text from Twitter is a separate and limited domain.

## Twitter Challenges

The language of Twitter is, of course, not always an advantage. It presents many unique challenges. People are less concerned about correctness (Metcalfe 2010), and they tend to be very creative when composing tweets.

### Morphology and Lexicon

Tokens in Twitter are characterized by coined words and spelling variations of high-frequency words sometimes to minimize typing but also simply to vary from the norm (e.g. spelling 'months' as 'monthz'). In all such cases there is an expectation that readers will understand the variation. Invariably that is true. Aside from the systematic changes, spelling

9

errors are common. The tweet in Example 7 demonstrates several issues of spacing, capitalization, and punctuation and contains challenges around spelling and lexical understanding.

(7) Cake?@Username K gotta get the hair,eyebrows n nails done today b4 9 then I WISH I had sum one I could go CAKE wit :( sigh

Garden variety spelling errors can be handled by spelling correction algorithms. Other variations are more regular and can be learned by statistically based systems or itemized for rules-based approaches. An open question is to what extent the variations are an ongoing process of language change in the medium. We have observed several types of variations that exist today.

**Repeated letters**. Repeated letters commonly occur at the ends of words as in *hmmmmm* although a final position is not required. It is equally likely to happen within a token usually with vowels but this is not required either. Twitter processing tools will also have to deal with repeating syllables in examples like *hahahahaha*.

**Changed and dropped letters**. In the same way that the letter 's' can be replaced with a 'z', the digit '0' can be used instead of the letter 'o'. Common suffixes appear with missing letters as in *runnin* or *playd* and certain vowels can be dropped out as seen in example 10, which also demonstrates several of the other issues mentioned here.

**Homophones**. Shorter syllables are substituted for longer ones when they are phonetically identical as in *sumthing*. Homophonic numbers are used the same way as in *b4* for *before*. Or the number 2 in *2morrow*, which can also replace whole words as either *too* or *to*.

**Punctuation**. The problem of punctuation in tweets is one of both too much and too little compared to more mundane corpora. Complete clauses are run together, sometimes with no punctuation at all. Conversely, the rhetorical style of Twitter includes phrase endings as long runs of end punctuation often with question marks and exclamation points interspersed. Another interesting rhetorical device used in Twitter is bracketing phrases in asterisks. Phrases bracketed in this way are presented as stage directions to dramatize the content, for example:

(8) *Thinks he may change his jeans.....stands looking in the mirror* does my bottom look big in these?

Asterisks are also used to emphasize words and as letter substitutions in expletives presumably to lessen any offense.

Other peculiarities stem from the conventions used in and the nature of the medium itself. Hash tags are designed to be used as keyword tags. However, there are many instances where the hash tag is incorporated into the content of the tweet.

(9) who wants to do my #mathhomework?

Parsing is further complicated by the fact that whole phrases can be embedded within hash tags.

(10) #IHateWhen I tell ppl I LOST SUMTHNG nd the 1st thng they ask me is WHERE U LOST IT AT? smh, if I knu i wouldn't b telln u I LOST IT @DUMMY

Username tags are likewise used creatively, but treating them as proper nouns covers most circumstances.

### Cultural and Topical Content

The content of Twitter tends to be topical and deeply entrenched within the culture of the people tweeting and their followers. Any system that hopes to understand Twitter input has to cope with current and trending background knowledge. Example 11 is a very simple instance of a current, cultural reference.

(11) LOL "Is there any room in this pocket for a little spare Chang?" #Community

The tweet refers to the television show *Community* and is making a play on words between 'change' and a character on the show whose name is Chang.

### Ellipsis

Ellipsis, the act of leaving out words and phrases, is a place where the shortness of the input works against language understanding instead of helping it. Twitter is rife with examples of both syntactic and semantic ellipsis.

(12) Want some chocolate

In example 12 the missing syntactic elements make this ambiguous as to whether it is asking a question or expressing a desire. Handling ellipsis is a difficult problem in NLP and arriving at some solution will be necessary to handle Twitter data.

The topical challenges and issues of ellipsis are difficult to solve but they only surface when attempting a deep understanding of Twitter text. The morphological and lexical issues as well as syntactic challenges are reasonable to solve. However, because of their peculiarities, current, off-the-shelf language processing tools do not work very well on Twitter text.

## Our Approach

We decided to tackle these issues by building a tool chain designed specifically for Twitter input strings. We created a tokenizer, a customized lexicon, and a parser to deal specifically with Twitter text. Recent, previous work indicates that others are interested in similar access to informal online content and have considered some other approaches. In (Foster 2010), the author evaluates the performance of different parsers on the informal text of blogs, wikis, and online discussion groups. This study attempts to transform the input text making it more like the training text for existing parsing tools. It also considers transforming training data to make it resemble the kind of informal text commonly found online. The project described in (Eisenstein et al. 2010) is, like us, interested in Twitter data specifically. The researchers consider regional variations in the language used by people in different parts of the United States. They leverage additional metadata (geotagged tweets) to consider new techniques for studying the language and even making predictions about speakers based on their regional distinctions. The recent literature contains many examples of people tackling the problem of gaining a deeper understanding of online text using various, interesting approaches. Our tool chain is described in the following sections.

**Tokenizer**

Existing tokenizers actually perform reasonably well on Twitter; however, we were able to create some efficiencies by handling Twitter-specific quirks at this level. In some cases post-processing output from a tokenizer would be difficult; for example, distinguishing semi-colons in emoticons from end-of-phrase punctuation, a problem our tokenizer can deal with in its single pass over the tweet string.

One of the first decisions we made was to collapse adjacent repeating characters into a single instance of that character, so, for example *pleaaaaaase* becomes *please*. This solves two problems. Repeating characters are extremely common in tweets both for effect and apparently because of 'finger bounces' when entering text. Equally common are dropped characters especially when they would normally repeat as in a word like *feel*. By eliminating *all* repeating characters we solve the problem of determining when to collapse and when not to, and we add robustness in dealing with two common types of spelling errors. The technique does, of course, introduce a new problem in that it changes legitimate spellings where a single letter is repeated, rendering a tokenized form of *speling* for example where the lexicon would have *spelling*. This is easily handled by processing the lexicon to convert each of the tokens to its revised spelling. It creates an additional complication because a word like *bee* is now listed as *be*, which of course collides with the entry for the infinitive form of the copula. We simply add a noun part-of-speech tag to the list of POS tags for the entry. (This is actually handled automatically when we generate our lexicon as described in the next section.) Since resolving multiple POS tags for a single token is already something the parser has to deal with, the additional effort incurred by this decision is minimal.

Similarly, apostrophes are removed from contractions, so that *we're* becomes *were*. Again, this creates an ambiguity between the contraction and the copula, but this change can also be resolved with little extra effort in the parser. Long runs of end punctuation (periods, question marks, and exclamation points) are also collapsed into a single instance. In this case no ambiguity is introduced although we do mark the tweet as containing repeated punctuation for whatever semantic value that might have.

Conventional tokenizers lose the fact that the @ symbol may or may not have a following space, which is important because the symbol can be used to mark a username tag or as shorthand for the word *at*. Other spacing in tweets can also be unusual:

(13) wr....ermm...NOT
   early.isnt
   note,credit

Our tokenizer is designed to handle these issues in a sensible manner.

Creating our own tokenizer also allowed us to handle emoticons during the initial processing of the input string. Emoticons are normally meant to represent a face and an emotional expression on the face. Our tokenizer includes a recursive transition network (RTN) designed specifically for processing emoticons.

There are many types and variations of emoticons:

| | | | | |
|---|---|---|---|---|
| :/ | :/ | ;-> | :-> | :D |
| :) | ;p | :P | :S | :0 |
| :L | :/ | : | D: | :o |
| :O | :S | :[ | ]: | <3 |
| =) | (= | :=) | (=: | =S |

There are many more besides the ones shown here, but there is a limited set of characters that can serve as the eyes and as the mouth of the face. For example, eyes are mostly created using a colon or semi-colon. Depending on the directionality of the emoticon :) or (:, a mouth or eyes character triggers a push into the emoticon RTN. There are also variations that are not meant to represent faces such as a heart (<3).

URL's, hash tags, and username tags are also treated as first-class entities by the tokenizer, which must also deal with the fact that as mentioned '@' can be used for the word *at*, and '#' can be used to mark a numeral as in *we're #1*.

**Lexicon**

Our lexicon is comprised of three separate lists of words. Starting from a word list containing approximately 92,000 words and their sets of possible lexical categories, we performed the transformations discussed above in the section describing the morphology and lexicon of Twitter. We also manually created a Twitter specific list for words that commonly appear in Twitter but are not in the conventional word list and assigned them lexical categories as well. These words and their parts of speech (POS) were used as an overlay to the starting list so that we could also use it to override POS tags where they differ in Twitter. For example, the word *peeps* (people) does not appear in the original word list, but is found often in Twitter. We added it to the list as *peps* according to our tokenizing rules and identified it as a plural noun. The word *kindle* exists in the original list but only as a verb. We added it to the Twitter-specific list identifying it as both a verb and a proper noun to override the original POS list.

The third word list came from a collection of approximately 92,000 people's names. This list was merged with the other two. When merging, if the name matched an item already in the lexicon, the tag for a proper noun was added to its set of lexical categories. If the name did not already exist, it was added to the lexicon and identified simply as a proper noun.

Collapsing repeating adjacent letters, caused collisions between some entries. After our transformations the words *ten* and *teen* become a single entry *ten* with the union of their sets of POS tags, so that the entry identifies the token as a cardinal, a noun, and an adjective. The parser then resolves the correct tag at parse-time.

Because of the irregular use of apostrophes in contractions, they are removed during morphological analysis. We created entries for them as such in the lexicon. We also created new POS tags for them to reflect their actual roles in the sentence, so that the contraction *I'm* becomes *im* and we identify it as a PPBE indicating that it is a combination of a personal pronoun and a form of *to be*. Interestingly, *im* is also an abbreviated form of *instant message*, so in addition to PPBE it is also marked as a noun and a verb.

**Parser**

The Twitter parser and grammar we created perform the minimum processing necessary to locate questions among Tweets. We have not yet created a full parser for Twitter although extending our current parser would not be too difficult. The parser is implemented using context-free rewrite rules and a simple top-down algorithm although we will most likely employ a more sophisticated approach for future work. Our primary goal at this point in the study was to determine the extent to which we could succeed in parsing tweets.

We approached the question-finding task as a search problem, where the grammar provides options for a list of possible parses all of which are formed as questions. The parser considers the different ways the constituents can combine to form a question as defined in the grammar. Searching is performed depth-first where various rule expansions are pushed onto a stack of possibilities. The current state consisting of the word position and the symbol list from the top of the stack are matched against the input sentence.

Our grammar contains 500 rewrite rules. Questions are organized according to and identified by type: wh-questions, auxiliary-initial, *be*-initial, and tag questions.

As mentioned above apostrophes are used inconsistently in tweets, so rather than parse a question word like *what's* into the two tokens *what* and *'s* as is commonly done, we handle such constructions lexically. The following list shows some interesting variations on wh- contractions from our lexicon.

| hows | wats | whatre | wheres | whtcha |
|------|------|--------|--------|--------|
| wasup | wazat | whats | whos | wtf |
| watcha | whatare | whatz | whose | |

Tweets often begin with any of several markings that are extraneous to understanding the syntax. The parser skips over initial user name and hash tags, URL's, emoticons, and other punctuation. Our grammar contains entries for initial interjections and vocatives so these are handled properly as well.

For an example of our grammar, consider a common class of questions formed by starting a tweet with 'anyone' or one of its variations (any1, anybody, anything). We created a classification for question pronouns, which we tag as *PPQ*. There are several grammar rules starting with PPQ to capture this class of questions. An input sentence like *any1 wanna talk* is captured by the rule:

```
PPQ MD VB
```

Much of the Twitter-specific language is handled lexically. Conventions like 'ru' for the phrase 'are you,' are treated as lexical items, so the rule

```
BEPRP VBG
```

parses tweets like *ru listening* where the tag BEPRP identifies the contraction of 'are' and 'you'.

**Results**

Our original and primary study (S.A. Paul and Chi 2011a) was to determine the Q&A behavior of the general population of social network users. Secondarily we wanted to know how predictive well-formed questions are of detecting information-seeking questions. We obtained candidate tweets and used Amazon's Mechanical Turk service to classify those that are questions and those that are not. To obtain candidate tweets we collected a random sample of 1.2 million tweets from the public Twitter stream. We then filtered these tweets by removing non-English, retweets and tweets that did not contain a question mark. Certainly there are question tweets that do not contain question marks, but our goal was to differentiate types of questions, so limiting the sample to those that are likely questions helped to obtain a manageable amount of data to work with.

We then designed a human-intelligence task (HIT) on Mechanical Turk in which we presented Turkers with candidate tweets and asked them to classify them. Others have used Mechanical Turk to rate computer-generated questions (Finin et al. 2010). They found a high correlation (74%) between Turkers ratings and those of a computational linguist serving as an independent judge. Although they looked at computer-generated questions, this may indicate that crowdsourcing the categorization of tweets can work successfully.

Our HIT contained the following instructions:

> Please read each of the following tweets and tell us whether you think this tweet is a question posed by the Twitter user to his/her followers with the expectation of getting a response.

Turkers were asked to classify 25 tweets as question, not a question, or not sure. Each candidate tweet was rated by two Turkers. If both Turkers rated a tweet as a question, we then classified the tweet as a true question.

Since workers on Mechanical Turk have been known to spam (Kittur, Chi, and Suh 2008), we designed several controls to ensure the validity of the data collected. First, Turkers were required to be Twitter users; this helped ensure that they were familiar with the language of Twitter and could understand the tweets in order to classify them correctly. Before they could do the HIT, Turkers were asked to enter their Twitter username, which was then verified with the Twitter service.

Moreover, to deal with the problem of spam responses, we inserted some control tweets along with the candidate tweets. Control tweets were tweets that we deemed easy to understand and were obviously questions or not questions. Each HIT consisted of 25 tweets; 20 candidate tweets and 5 control tweets. In our subsequent analysis, we only included data from those Turkers who rated all control tweets correctly.

For our parsing test, we processed 2304 tweets from our set of labeled tweets where 1152 of the tweets were deemed to be questions and 1152 were deemed not to be questions by Turkers. After processing the tweets with our parser, we obtained the following counts of tweets where the Mechanical Turkers and the parser agreed on both questions and non-questions and where the two differed.

|  | MT Question | MT ¬Question |
| --- | --- | --- |
| Parser Question | 898 | 486 |
| Parser ¬Question | 254 | 666 |

Yielding the following scores: **Precision**: .64884 **Recall**: .77951 **Accuracy**: .67881

## Conclusion

Our scores are rather low and indicate that well-formedness is probably not a good indicator of information-seeking questions on Twitter. This is not completely surprising considering that the example questions (Examples 1, 2, and 3) illustrate that question syntax is used for a variety of reasons and not just for finding information. However, apart from the disappointing scores, our more casual observations of the performance of the parser on the task of finding tweets in the form of syntactic questions indicate that it was able to find syntactically formed questions very well. We did not test this tertiary question and have concluded that simply knowing the syntax of a tweet is not interesting enough by itself. Most studies will require a deeper understanding of the content, although we still believe that the syntax can be used as a feature in a broader machine learning system to detect more interesting aspects of tweets. We are continuing with that work now.

The larger question as to whether or not deep parsing is possible and tractable on Twitter data has not been answered yet. It will, however, most likely depend on an ability to understand the syntax of tweets and our experience with that has been encouraging. Other projects such as (Gimpel et al. 2011) are producing additional useful data that can help the community develop richer tools leading to deeper automated understanding of Twitter and online language.